# Toward an example-based machine translation from written text to ASL using virtual agent animation

Mehrez Boulares[1] and Mohamed Jemni[2]

Research Lab. UTIC, University of Tunis
5, Avenue Taha Hussein, B. P. : 56, Bab Menara, 1008 Tunis, Tunisia

**Abstract**
Modern computational linguistic software cannot produce important aspects of sign language translation. Using some researches we deduce that the majority of automatic sign language translation systems ignore many aspects when they generate animation; therefore the interpretation lost the truth information meaning. This problem is due to sign language consideration as a derivative language [9, 15], but it is a complete language with its own unique grammar [1, 2]. This grammar is related to semantic-cognitive [9, 18] models of spatially, time, action and facial expression to represent complex information to make sign interpretation more efficiently, smooth, expressive and natural-looking human gestures. All this aspects give us useful insights into the design principles that have evolved in natural communication between people.
In this work we are interested in American Sign Language, because it is the simplest and most standardized sign language [2]. Our goals are: to translate written text from any language to ASL animation; to model maximum raw information using machine learning and computational techniques; and to produce a more adapted and expressive form to natural looking and understandable ASL animations.
Our methods include linguistic annotation of initial text and semantic orientation to generate the facial expression. We use genetic algorithms [6, 14, 19] coupled to learning/recognized systems to produce the most natural form. To detect emotion we based on fuzzy logic [20] to produce the degree of interpolation between facial expressions. Roughly, we present a new expressive language Text Adapted Sign Modeling Language TASML that describes all maximum aspects related to a good sign language interpretation.
This paper is organized as follow: the next section is devoted to present the comprehension effect of using Space/Time/SVO form in ASL animation based on experimentation. In section 3, we describe our technical considerations. We present the general approach we adopted to develop our tool in section 4. Finally, we give some perspectives and future works.

## 1. Introduction

According to many studies, reading levels of hearing impaired is lower than the reading level of hearing student. Moreover, in 1996 Marschark and Harris confirmed that their learning progress is extremely slow. Furthermore, the gain of experience collected by deaf children in four years is equivalent to the gain of one year for hearing children.

Many deaf people in the world have difficulties with reading and writing; in fact some of them do not read. A text-based system would make it impossible for the people to follow and understand a conversation in real-time. If spoken language is rendered as text, all the information related to the semantic-cognitive meaning, time, space, emotion, intensity of the signs and facial expression is lost [2, 11].

Using ICT we can improve the classic pedagogical methods by new e-learning methods [7, 8, 12] based on multi-media contents [10, 16] and sign language specificities for a better discourse interpretation. Research in computational linguistics, computer graphics and machine learning systems has led to the development of increasingly sophisticated system based on virtual character over the past few years, bringing new perspective to machine translation research [3].

As the primary communication means used by members of deaf community, sign language is not a derivative language. It is a complete language with its own unique grammar [1, 13]. But the majority of automatic interpretation system ignores this important aspect therefore the interpretation lost the truth information meaning. In our study, we considered the American Sign Language as a target and any textual language as an input. We use the Google translator to translate all input language text to be entered in English before starting.

In this paper, we present a new approach developed in our Research Laboratory which can provide benefits for deaf people with low English literacy to improve their social integration and communication. As a first step we focus on Sign Language specificities to generate automatically the text description to more adaptive form represented in new descriptive language Text Adaptive Sign Modeling Language TASML. This new approach uses mainly a web-based interpreter of sign language developed in our research laboratory and called WebSign [7]. It is a tool that permits to interpret automatically written texts in visual-gestured-spatial language using avatar technology.

## 2. Comprehension effect of the use of Space/Time/Subject/Action form in sign language:

In order to determine the more adequate textual form to be interpreted in sign language. We conducted experiments with Sign Language animations in which some animations are conformed to the place time subject action form and other versions of the animations have these linguistic phenomena removed. For experimentation, we have a specialist sign language interpreter which interprets the news program and some deaf participants. Before discussing this latest study, we will first briefly survey some earlier experiments about general specificities of sign language. The methods and results of this earlier study served as a foundation for the design of our most recent study.

2.1 Earlier work on sign language specificities:

In order for us to conduct studies in which we focus on specific linguistic phenomena, we first needed to ensure that our animations have a baseline level of understandability. To this end, we needed to prove the existence of phenomena models representing sign language specificities. Furthermore, Patrice dale gives us an idea about signing space model for the interpretation of sign language interactions. Also the experimental study of Matt Huenerfauth [5] to improving spatially references, given us the following results:
Participants in the study viewed 16 ASL stories of two versions: "space" and "no-pace" animations. The average comprehension question scores for the space animation are greater than no space animations. This explains the importance of spatial reference in discourse interpretation. As other aspects we find speed/timing [4] configuration and iconicity in sign language. Model of the speed, timing and iconicity [2, 18] of ASL animations is very important to produce more understandable animation for better text interpretation [17].

2.2 Current experiments evaluating use of Place/Time/Subject/Action form with Emotion facial expression:

In our laboratory, we conducted a new study to determine whether there is a comprehension benefit when Sign Language animations include signs that organize the item that represent place reference at first position, time reference at second, subject at third and finally the action at fourth position.

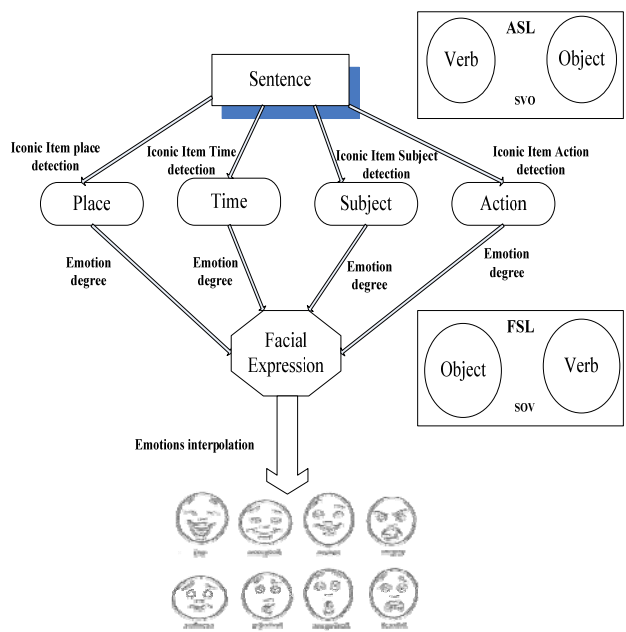

Fig. 1: Sign Language Linguistic form

We designed for explication, 7 sentences in American Sign Language and French Sign Language, and our interpreter translates these sentences to 7 deaf participants. The first version was produced in the original text order without facial animation to describe Emotion. The second version of each sentence was produced in our abstract form using facial expression to describe emotion in every sentence. If sign language target is the FSL we respect the grammar related to SOV (Subject Object Verb) [2, 15] form and if animation result is in ASL we respect the SVO (Subject Verb Object) [9, 15] form. The action items are followed or preceded by Object according to target sign language.

---

**Original Text Order**

1- The Viking period in history stretched for about three to four hundred years from 790 after Jesus-Christ to 1100 after Jesus-Christ.
2- During this period, Viking warriors raided nearby lands, explored uncharted seas, and searched for and found trade routes throughout Britain, Ireland, Southern Europe, North Africa, and Central Asia.
3- Vikings or Norsemen originated from what are now Norway, Sweden, and Denmark.
4- The geographical location of these three Norse countries determined the routes their inhabitants took to raid, explore, and trade.
5- The first recorded raids occurred in 793 after Jesus-Christ. When warriors raided a small island on the northeast coast of Britain and attacked the monastery on Lindisfarne.

6- The last recorded battle took place in 1066 when the Norwegian king, Harald Hardrada, invaded England to claim the throne of Edward the Confessor.
7- The new Anglo-Saxon king, Harold Godwinson defeated Harald in Yorkshire on September 25.

---

1- The Viking period in history stretched for about three to four hundred years from 790 after Jesus-Christ to 1100 after Jesus-Christ.

### ASL-SVO Order

**Place:** [null]
**Time:** [Past: for about three to four hundred years from 790 to 1100 after Jesus-Christ [**pause: 1Second**] à 1100]
**Subject:** [The Viking period in history] [**pause: 0.5Second**]
**Verb:** [to stretch]
**Object:** [null]
**Emotion:** [Surprise] [Fear] [Surprise-Fear]
**Finger Spelling:** Name
**Speed:** 1 Sign/Second

---

2- Au cours de cette période, les guerriers viking envahissent les terres avoisinantes, ils explorent des mers inconnues, cherchent et trouvent des routes commerciales à travers la Grande-Bretagne, Irlande, Europe du Sud, l'Afrique du Nord et en Asie centrale.

### FSL-SOV Order

**Lieu :** [les terres avoisinantes]
**Temps :** [Passé : au cours de cette période] [**pause : 1Seconde**]
**Sujet :** [les guerriers viking]
**Objet :** [les terres avoisinantes] [**pause : 1Seconde**]

**Verbe :** [envahir]

**Objet :** [des mers inconnues] [**pause : 1Seconde**]
**Verbe :** [explorer]
**Objet :** [des routes commerciales [**pause : 1Seconde**] à travers la Grande- Bretagne, Irlande, Europe du Sud, l'Afrique du Nord et en Asie centrale.]
**Verbe :** [chercher et trouver]
**Emotion:** [Colère] [surprise] [Colère-surprise]
**Finger Spelling :** Noms propres
**Vitesse :** 1 Signe/Seconde

---

Fig. 2: Transcription of three different versions of the same sentences

Fig. 2 contains transcripts of three different versions of same sentences here we put only some sentences to just explain the output form to be interpreted. In original text we use only the input text order to be interpreted in this way. Respectively in SVO and SOV form the interpreter translates the sentences according to the pivot language related to each sign language with emotion, sign speed [4], and finger spelling for nouns. According to Matt Huenerfauth study of the calculating linguistically motivated durations for the signs in an ASL animation, we put some pauses between signs and sign speed is parameterized at 1 Sign/Second. Our sign language expert interprets first the original text and then each one of 7 participants describes in his own way what he has understood. We repeat this operation for ASL and FSL [2] form next we compare the results. We used the two sign language ASL and FSL, to prove that our form is not related to some specific sign language but is a more general approach

As results of the study, Figure 3, 4 and 5 displays the average comprehension scores for participants in the study for the Pivot/Emotion vs. no- Pivot/Emotion. As shown in Fig. there was a statistically significant difference in Understandability scores between the Abstract/Emotion order and no-Abstract/Emotion version of the animations with considering some other parameters like pauses delay and sign speed [4]. Figures 3, 4 and 5 shows the participants average interpretation scores for the 1-to-7 participant scale.

Each participant interprets the sentence in his own way and our expert gives us an approximate percentage value in understandability/Grammaticality/Naturalness [2]. The differences between the Pivot/Emotion and no-Pivot/Emotion version is very clear. Participants give higher scores for the Pivot grammatical order, understandability, and naturalness of this structure according to the animations. The comprehension benefit observed in this study is related to the use of the more appropriate text structure and emotion when translating into sign language which facilitates the participants to understanding information in the story;

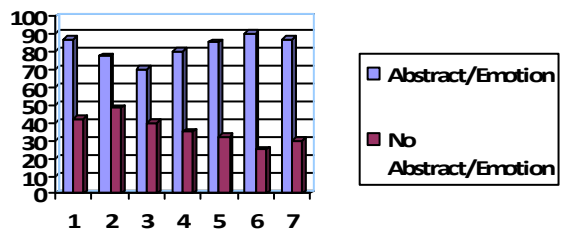

Fig. 3: Understandability

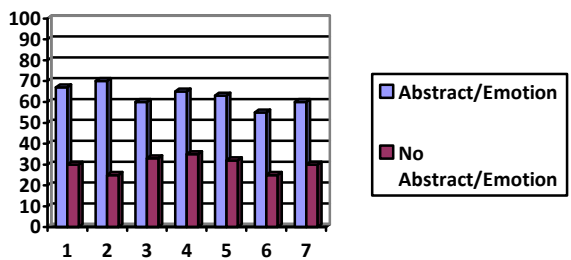

Fig. 4: Grammaticality

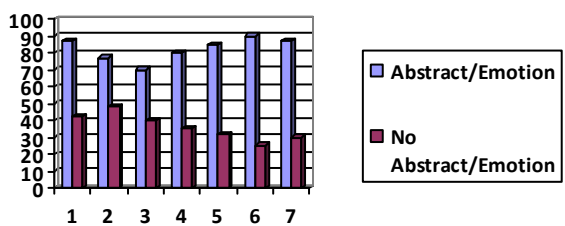

Fig. 5: Naturalness

## 3. Technical considerations

The objective of our project is to develop a Web based interpreter of Sign Language (SL). This tool would enable people who do not know SL to communicate with deaf individuals. Therefore, contribute in reducing the language barrier between deaf and hearing people.

Our secondary objective is to distribute this tool on a non-profit basis to educators, students, users, and researchers and to encourage its wide use by different communities. The developing team has adopted Web 3D technologies [7] which have surpassed traditional approaches such as video and still images. We use the Sign Modeling Language SML as transcript language to animate our 3D humanoid interpreter.

SML is based on XML structure to describe animation such as X3D format [7]. Actually our system is based on automatic translation from the written text without considering the sign language specificities such as the cognitive form according to the sign construction, facial expression, timing and speed parameters. In this work we focus on the transformation of initial text to more adapted form related to target sign language. Furthermore, to aim more comprehensive animation produced by our system and to represent a general text form whatever the initial text language and the target sign language which can be used by other transcription systems.

### 3.1 Translation system architecture

Our transcription system architecture is based on sign modeling language data base. The system sends the input text as a request to Google translator service. This service give' us English translation to be an input to the Text Adapted Sign Modeling Language module. After this operation we obtain a text form adapted to the sign language specificities and which represent an input to WebSign kernel [7]. Finally this engine generates the sign modeling language SML [7] according to this text form to be interpreted by our virtual character (Figure 6).
At first, the system support all input text language according to Google translator but the result will be in American Sign Language ASL of course the system can be parameterized to support other sign language

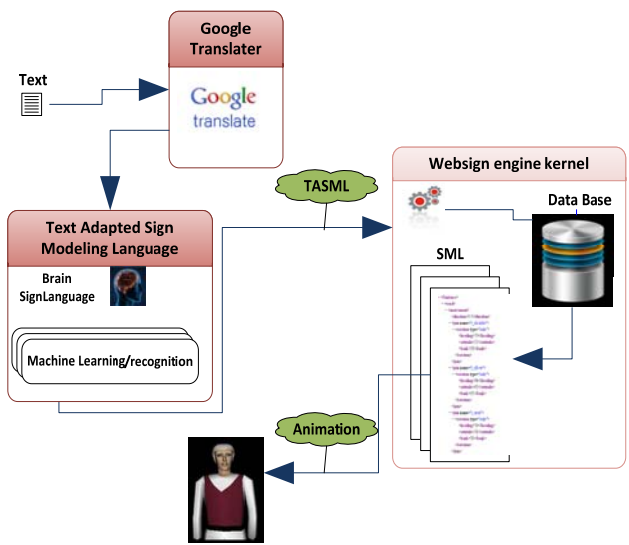

Fig. 6: Translation system architecture

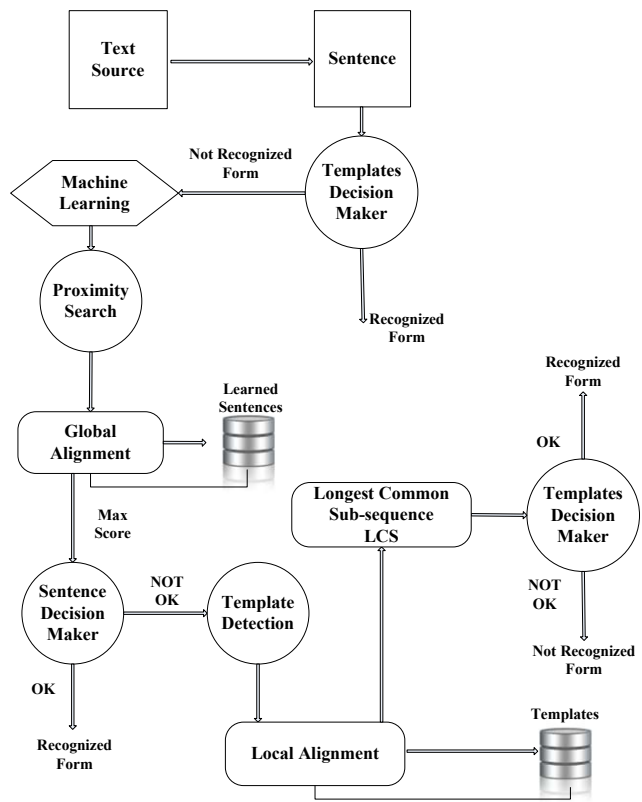

Fig. 7: Global system architecture

## 4. Our approach

Based on machine learning systems, biological algorithms and fuzzy logic concept. We have built algorithms for generating the different elements that constitute our adapted text form and algorithms for emotion detection

### 4.1 Text abstract form generator module Acknowledgments

Our solution is based on Example-Based machine learning system associated to genetic algorithms. The aim is to use the power of genetic algorithms for detecting different sequences existing in the initial sentence. Furthermore, we proceed to fetch if the input sentence was learned by our system or no. If our system doesn't recognize this sentence, the proximity search is launched to get the better sentence proximity.

According to the threshold, the system makes decision about the degree of correctness recognized form. If this part of decision is not enough the template (Example) detection module takes over to detect different part that are used to learn our system.

#### 4.1.1 Global proximity search

We have implemented an algorithm for detecting the global proximity between the input sentence and the learned sentences. This part of system allows us to detect the more appropriate form according to input sentence. However, Bioinformatics is the solution for calculating the proximity degree between different sequences.

The genetic algorithms are created for comparing DNA and protein sequences and exploit it to calculate the global proximity degree between character sequences. The idea is to use the power of these algorithms through the search heuristic that is routinely used to generate useful solutions to optimization and search problems.

They are many algorithms for comparing different sequence, the better one is The Needleman–Wunsch algorithm [14] performs a global alignment on two sequences. It is commonly used in bioinformatics to align protein or nucleotide sequences. This algorithm belongs to the dynamic programming for solving complex problems by breaking them down into simpler sub problems, and was the first application of dynamic programming to biological sequence comparison.

We start first by splitting the paragraph into sentences, next we proceed to a global alignment between each learned sentence and the first sentence. After this step we calculate the max proximity score between all obtained score. The algorithm is described in Figure 8.

---

**Algorithm:** Proximity-Search

  **F**: Matrix [length (SEQ1) +1, length (SEQ2) +1]

  **Basis**:

   F0j = d * j

   Fi0 = d * i

   S (Ai, Bi) is the similarity of characters Ai and Bi

   Recursion, based on the principle of optimality:

  **Fij = max (Fi-1, j-1 + S (Ai, Bi), Fi, j-1 + d, Fi-1, j + d)**

  Score = F [length (SEQ1), length (SEQ2)]

  Compare (Max (scorei), threshold)

  If SentenceDecisionMaker = OK then

  Return TargetForm

  Return LearningStep

---

Fig. 8: Proximity Search Pseudo Code

In order to perform a Needleman-Wunsch alignment, a matrix is created which allows us to compare the two sequences. The score as determined through use of the above calculation is placed in the corresponding cell. This algorithm performs alignments with a time complexity of O (mn) and a space complexity of O (mn) [14].

### 4.1.2  Local template (Example) proximity search

Our algorithm tries to search first if the input sentence was learned or no. If the system doesn't recognize automatically the input sentence, the algorithm builds a global alignment to determine the closest sequence, if this sequence is very similar according to the threshold, the system gives similar output related to the founded sentence and this is how the system learns the new sentences.

Now, if the sentence is not recognized, also our algorithm try automatically to fetch all different templates (Examples) related to the abstract (Pivot) form to be clustered. We recall that our abstract form is composed by all the elements related to the subject place indication at first position, at second position we find the time parameter, at third position the subject takes his place in abstract form and finally the action can begin and all elements according to the action are clustered in this position.

In this step, the objective is to learn system how gives us the desired target form. However, we use the local proximity search and consequently the local alignment algorithm. This approach is related to the more appropriate algorithm between local and global alignment according to the context.

The global alignment considers all elements of a sequence. If the lengths of sequences are different, insertions / deletions are introduced in the shortest sequence for the two sequences are aligned along their entire length.

This algorithm is most useful when the sequences in the query set are similar and of roughly equal size. We use it at first step because we just need a few characters differences between the initial sentence and the learned sentence.

However, the local alignment looks between the two sequences to align the sequence fragments that align best. This kind of alignments is more useful for dissimilar sequences that are suspected to contain regions of similarity or similar sequence motifs within their larger sequence context.

The Smith-Waterman algorithm [19] is a general local alignment method also based on dynamic programming. However, is a well-known algorithm for performing local sequence alignment; this algorithm gives very good results when we need to determining similar regions between two nucleotide or protein sequences.

In our context we use this algorithm to determining the different elements that can be clustered according to proximity between the learned cluster (Example) and the input sentence. Furthermore, Instead of looking at the total sequence, the Smith–Waterman algorithm compares segments of all possible lengths and optimizes the similarity measure.

Using this approach, our recognizer requires a set of templates for matching the input sentence with all learned models. At the end of this operation we obtain a new sentence conformed to the place, time, subject, action form and this sentence will be a template for the next input text.

```
Algorithm: Pivot form recognition

S: Input sentence

WS: {word_1, word_2,…,word_n}, word ∈ S

TS: {Template_1, Template_2,…,Template_n}

T: Template, T ∈ TS

WT: {word_1, word_2,…,word_n}, word ∈ T

LS: Learned Sentence

For Template in TS

  For wordT in WT (Template)

    For wordS in WS

      Sc = Smith-Wat(wordT, wordS)    } Max(Sc)
    end.                                        } Max∑(Sc)
         ∑ ( Sc)
  end.

end.
```

$$LS = Max \left( \sum Smith\text{-}Waterman\ (WT, WS) \right)$$

If length (LCS (SL, S)) ≥ length(S) – threshold Then

    Return True

    Return False

Fig. 9: Pseudo code abstract (Pivot) form recognition

Figure 9 shows the pseudo code abstract form recognition from the input sentence using all learned templates. For each template word a local alignment applied with every word of the initial sentence. Then the system tries to find the maximum score alignment according to all proximity search operation. For each template word a local alignment applied with every word of the initial sentence. Then the system tries to find the maximum score alignment according to all proximity search operation. After this step we will have a set of score which represents the proximity between all words in the two sentences.

In order to output the best form related to the initial sentence, our algorithm repeats this operation and compares the added score of all template and choose the highest score. Therefore, the system gives us the most appropriate form to our initial sentence but this may be a wrong result. In order to ensure the learning process automatically, we implements an algorithm that finds the longest common subsequence LCS [6] between the source sentence and the result.

This step gives us very good results by comparing the LCS length and input sentence length to make decision to conserve this result or no. Certainly this phase requires a long execution time for all new sentences but after the recognition process of many sentences the system response time improves considerably due to the learning already done by machine learning system.

The LCS computation complexity, is evidently in O (m*n) time and space related to the matrix size for two sequences. The global alignment Needleman-Wunsch complexity is in O (m*n) time and pace also the local alignment Smith-Waterman is in O (m*n) [6].

### 4.1.3 Automatic detection of simple/compound emotion

Our approach in generating abstract (pivot) form related to the place, time, subject, action order detects also the different elements that reference to the emotion in sentence. But sometimes we find many emotions like surprised, feared and other in the same sentence. When the system interprets the text automatically, we lost the exact meaning related to the emotion and this is a big problem.

In order to find solution, we are based on the observation of the natural deaf interpretation for this phenomenon. With collaboration of our sign language specialist, we deduced that the sign language interpreter read first the text, he understand it and he translate the global text meaning. Therefore, when we focus on his interpretation we presume that some emotions are mixed automatically to produce the truth meaning.

According to studies of psychiatrist Jacques Cosnier expert in human and animal emotion a basic emotion classification is built figure.

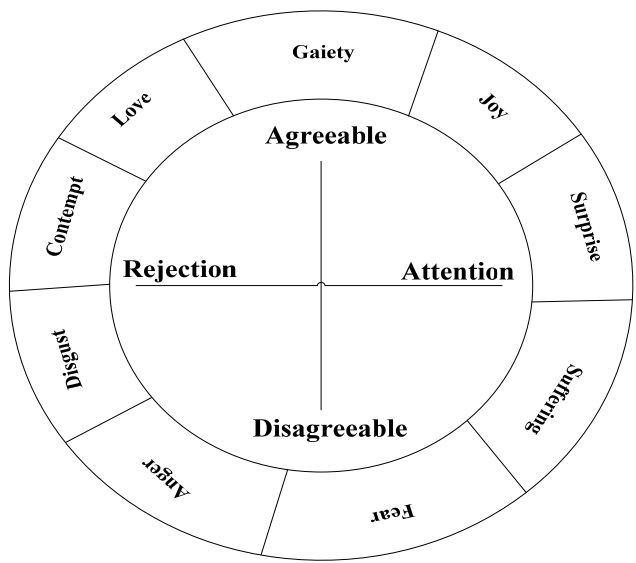

Fig. 10: Organizing axes emotions

This figure shows that there is some overlap between the basic emotions, for example it makes sense to find the emotion of joy and surprise, surprise and suffering, suffering and fear, fear and anger, anger and disgust, disgust and contempt, contempt and love, love and gaiety, gaiety and joy.

Therefore, this classification gives us the most appropriate emotion related to the context because in most of cases emotions are confused and mixed. We can find a mix between two or even three emotion in a single facial expression which gives us more precision.

For generating automatically the mixed emotions, we need a mathematical model to represent the desired results. For this reason we focus on the fuzzy logic which represents as best as possible our problem. Fuzzy logic emerged as a consequence of the 1965 proposal of fuzzy set theory by Lotfi Zadeh [20] though fuzzy logic has been applied to many fields, from control theory to artificial intelligence.

Fuzzy logic is a form of many-valued logic derived from fuzzy set theory to deal with reasoning that is fluid or approximate rather than fixed and exact. In our context we need to represent fuzzy information's related to the emotion classification. Furthermore, to make possible to detect one or more emotions in the same sentence. Consequently to make decision and to produce automatically the appropriate emotion.

In contrast with "classical logic", where binary sets have two-valued logic, if we use this binary logic the system produce wrong results according to the comprehension meaning. Fuzzy logic variables may have a truth value that ranges in degree between 0 and 1. In other meaning, fuzzy logic is a superset of conventional (Boolean) logic that has been extended to handle the concept of partial truth, where the truth value may range between completely true and completely false.

Consequently, we can represent the result with the emotion overlap criterion. In this theory we use the linguistic variables which are ideally suited to our problem. Furthermore, the problem may be managed by specific functions.

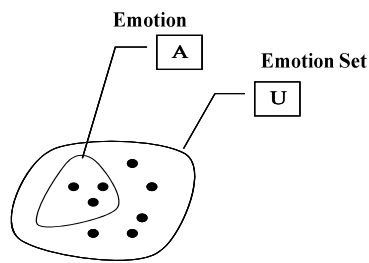

**Classical set theory:**

if µA is the membership function of the Emotion A set

$\forall x \in U$, µA(x) = Value1 if $x \notin$ Emotion A

µA(x) = Value2 if $x \in$ Emotion A

$x \in$ Value1 **or** Value2

If µA(x) = 0.4

x : is the interpolation value between two emotions

Value $_n$ : Emotion value

$x \in$ Value1

$x \notin$ Value2

**Fuzzy set concept:**

if µA is the membership function of the Emotion A set

$\forall x \in U$, µA(x) $\in$ [Value1,Value2]

If µA(x) = 0.4

x : is the interpolation value between two emotions

Value $_n$ : Emotion value

$x \in$ Value1 **and** Value2

$x \in$ Value1 at 40% and $\in$ Value2 at 60%

Fig. 11: Emotion fuzzy concept

This approach has yielded good results which will be useful in systems that use interpolation between the different facial expression models to produce emotion. For example: sentence which contains suffering and fear. Suffering intensity value detected from sentence is bigger than fear intensity value:

X $\in$ suffering at 70% and $\in$ fear at 30%, the result is an interpolation value between suffering and fear emotion which can be parameterized according to the target system.

## 5 Conclusion and future works

This research has identified a solution to improve the comprehension of Sign Language animations when using the automatic text interpretation into sign language. By modifying some parameters that have received little systematic research attention before: reorganizing the input text to an appropriate more adapted form based on surrounding sign language factors and automatic detection of simple and compound emotion.

This project is classified among the first studies to prove that deaf perceives a big difference in quality of Sign Language animations based on this specific form and automatic compound emotion detection from text. To confirm the theoretical approach of this contribution, we based on experimental support to shows that deaf perceive a difference in understandability, grammaticality and naturalness according to these specific form and emotion interpolation factors.

While this study focused on French Sign Language and American Sign Language, these techniques should be applicable to animations of other sign languages around the world. However, we just need to train system with a desired corpus language which contains the output templates (Examples) and sentences. Now that this study has shown that Sign Language animations can be made more understandable, our approach can be used in a real-system setting as part of a WebSign tool [7], we are going to integrate this system in WebSign to improve the animation quality.

In future work, we will focus on other sign language specificities to be learned by our system. We will study the space factor related to the objects location in discussion. The automatic management of the sign area is very important factor to give us a more significant animation closest to the natural animation. The promising initial results of this study open the question of whether there may be standardization of sign language when using a collaborative approach to generates all sign language corpus to be learned by this system. Therefore, an investigation approach based on data meaning can be applied to gives us the more appropriate standardized sign language.

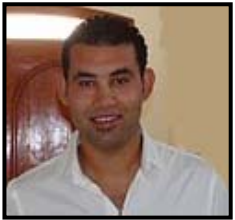

**Mehrez Boulares** is currently a PhD student under the supervision of Prof. Mohamed Jemni. He received in September 2009 the Master degree on Computer Science from Tunis College of Sciences and Techniques (ESSTT), University of Tunis in Tunisia. His research interests are in the areas of Sign Language Processing. His current topics of interests include Computer graphics and Accessibility of ICT to Persons with Disabilities.

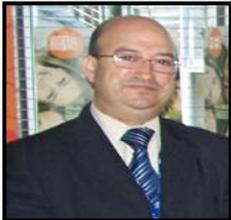

**Mohamed Jemni** is a Professor of ICT and Educational Technologies at the University of Tunis, Tunisia. He is the Head of the Laboratory Research of Technologies of Information and Communication (UTIC). Since August 2008, he is the General chair of the Computing Center El Khawarizmi, the internet services provider for the sector of the higher education and scientific research. His Research Projects Involvement are tools and environments of e-learning, Accessibility of ICT to Persons with Disabilities and Parallel & Grid Computing.